\documentclass[10pt,twocolumn,letterpaper]{article}

\usepackage{iccv}
\usepackage{times}
\usepackage{epsfig}
\usepackage{graphicx}
\usepackage{amsmath}
\usepackage{amssymb}
\usepackage{multirow}
\usepackage{algorithmicx,algorithm}
\usepackage[noend]{algpseudocode}
\usepackage[accsupp]{axessibility} 

\usepackage[breaklinks=true,bookmarks=false]{hyperref}

\iccvfinalcopy 

\def\iccvPaperID{****} 

\ificcvfinal\fi

\begin{document}

\title{Online Knowledge Distillation for Efficient Pose Estimation}

\author{
	Zheng Li \textsuperscript{\rm 1},
	Jingwen Ye \textsuperscript{\rm 2},
	Mingli Song \textsuperscript{\rm 2},
	Ying Huang \textsuperscript{\rm 1},
	Zhigeng Pan \textsuperscript{\rm 1}\thanks{Corresponding author}
	\\
	\textsuperscript{\rm 1} Hangzhou Normal University,
	\textsuperscript{\rm 2} Zhejiang University \\
	{\tt\small lizheng1@stu.hznu.edu.cn, \{yejingwen,brooksong\}@zju.edu.cn, \{yw52,zgpan\}@hznu.edu.cn}
}

\maketitle

\ificcvfinal\thispagestyle{empty}\fi

\begin{abstract}

Existing state-of-the-art human pose estimation methods require heavy computational resources for accurate predictions. One promising technique to obtain an accurate yet lightweight pose estimator is knowledge distillation, which distills the pose knowledge from a powerful teacher model to a less-parameterized student model. However, existing pose distillation works rely on a heavy pre-trained estimator to perform knowledge transfer and require a complex two-stage learning procedure.
In this work, we investigate a novel \textbf{O}nline \textbf{K}nowledge \textbf{D}istillation framework by distilling \textbf{H}uman \textbf{P}ose structure knowledge in a one-stage manner to guarantee the distillation efficiency, termed \textbf{OKDHP}. Specifically, OKDHP trains a single multi-branch network and acquires the predicted heatmaps from each, which are then assembled by a Feature Aggregation Unit (FAU) as the target heatmaps to teach each branch in reverse.
Instead of simply averaging the heatmaps, FAU which consists of multiple parallel transformations with different receptive fields, leverages the multi-scale information, thus obtains target heatmaps with higher-quality. Specifically, the pixel-wise Kullback-Leibler (KL) divergence is utilized to minimize the discrepancy between the target heatmaps and the predicted ones, which enables the student network to learn the implicit keypoint relationship. Besides, an unbalanced OKDHP scheme is introduced to customize the student networks with different compression rates. The effectiveness of our approach is demonstrated by extensive experiments on two common benchmark datasets, MPII and COCO.

\end{abstract}

\section{Introduction}

Human pose estimation aims to recognize and localize all the human anatomical keypoints in a single RGB image. It's a fundamental technique for high-level vision tasks, such as action recognition~\cite{cheron2015p}, virtual reality~\cite{pham2018human} and human-computer interaction. Since the invention of DeepPose~\cite{toshev2014deeppose}, deep neural networks have been the dominant solution for human pose estimation, based on which, the approaches~\cite{wei2016convolutional,yang2017learning,sun2019deep} focus on exploiting richer representations with a sequential architecture and achieve state-of-the-art performance. 
However, the gains of such deep learning based approaches often come with a cost of training and deploying the over-parameterized models, which limits the deployment in resource-intensive mobile devices.
To reduce the computation cost and enhance the model efficiency, many efforts have been devoted to directly designing lightweight and real-time networks, e.g., PAF~\cite{cao2017realtime}, VNect~\cite{mehta2017vnect} and MultiPoseNet~\cite{kocabas2018multiposenet}. 

\begin{figure}
	\centering
	\includegraphics[width=1\linewidth]{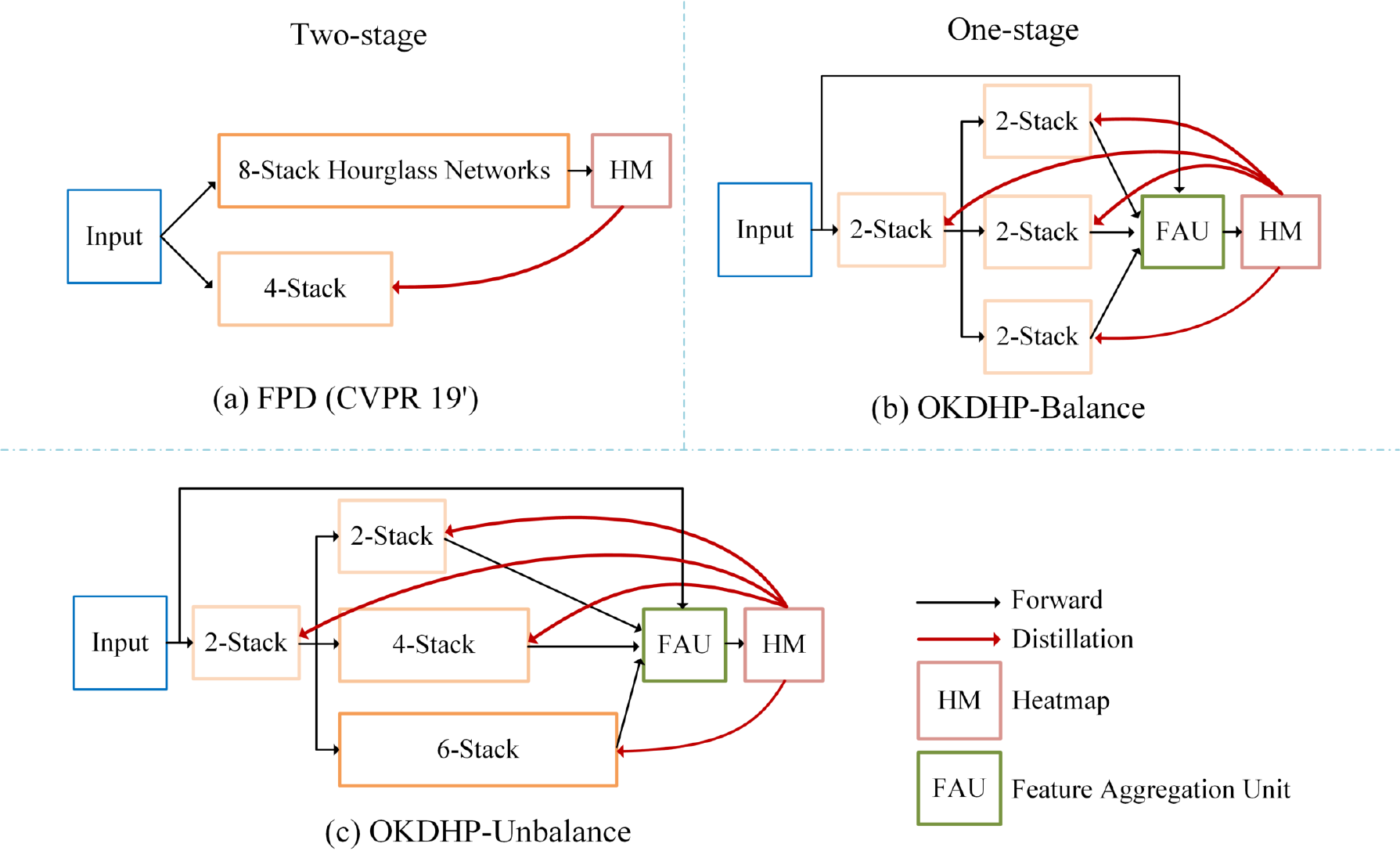} 
	\caption{
		To obtain an efficient 4-Stack network, (a) FPD~\cite{zhang2019fast} adopts a two-stage distillation scheme from the static pre-trained 8-Stack network. The proposed OKDHP distills the pose structural knowledge with both (b) Balance and (c) Unbalance architectures in one stage. The teacher is established online with the FAU.
	}
	\label{fig:okdhpv2}
\end{figure}

As another powerful tool to achieve a good trade-off between speed and accuracy, knowledge distillation~\cite{hinton2015distilling} follows the teacher-student paradigm. Traditional distillation utilizes a two-stage scheme that starts with a cumbersome pre-trained teacher model, then distills the knowledge to a compact student model.
In the field of pose estimation, recent works~\cite{zhang2019fast,hwang2020lightweight} adopt a traditional offline distillation scheme which distills the knowledge from a large pre-trained pose estimator (teacher) to a lightweight pose estimator (student) as shown in Fig.~\ref{fig:okdhpv2}(a).
However, training such a heavy teacher model is time-consuming and a high-capacity model is not always available.
Thus, online counterparts~\cite{zhang2018deep,zhu2018knowledge} are proposed to simplify the distillation process to one stage, reducing the demand for the pre-trained teacher model. In ONE~\cite{zhu2018knowledge}, a strong teacher model is established on-the-fly and all students share the same target distribution by averaging the predictions of all branches with learnable weights. Prior impressive works are mostly devoted to classification tasks, which neglect the valuable structural knowledge in the pixel-level tasks. Thus, our work focuses on the more challenging pixel-level tasks and proposes the first online pose distillation framework.

Existing pixel-level distillation works~\cite{zhang2019fast,hwang2020lightweight} use mean squared error (MSE) as the distillation loss which is weak for knowledge transfer. It can not effectively measure the relative entropy between two probability distributions. Besides, MSE is used as the loss function of both task-specific supervised term and distillation term. These two loss terms have different optimization targets, one is the ground truth heatmap, the other is the predicted heatmap generated by the teacher. The conflict between two loss terms will deviate the optimization into a sub-optimal situation.

To alleviate those limitations, we investigate an online pose distillation approach for \emph{efficient} pose estimation.
The proposed method has two vital aspects for efficiency. \emph{One is that we simplify the distillation procedure to one stage. The other one is that the proposed method significantly improves the pose estimation accuracy comparing with the original network.} The whole framework is constructed with a Feature Aggregation Unit (FAU) and multiple auxiliary branches, where each branch is treated as a student. The student branch can be both the same or heterogeneous architectures, making up the OKDHP-Balance and the OKDHP-Unbalance architectures respectively and enabling the customization of different compression rates. The teacher is established by the weighted ensemble of the predictions of all students through the FAU.
The FAU here captures the multi-scale information to obtain the higher-quality target heatmap.

Besides, to transfer the pose structural knowledge, the pixel-wise KL divergence loss is utilized to minimize the discrepancy between the target heatmaps and the predicted ones. In the final deployment, the target single-branch network is acquired by simply removing redundant auxiliary branches from the trained multi-branch network, which doesn't introduce any test-time cost increase. 

The main contributions of this paper are listed below.
\begin{itemize}
	\item To our best knowledge, we are the first to propose the online pose distillation approach, which distills the pose structure knowledge in one-stage manner. 
	\item Both balanced and unbalanced versions of OKDHP are introduced, which can customize the target network with different compressing rates.
	\item Extensive experiments validate the effectiveness of our proposed method on two popular benchmark datasets: MPII and COCO.
\end{itemize}

\begin{figure*}
	\centering
	\includegraphics[width=1\linewidth]{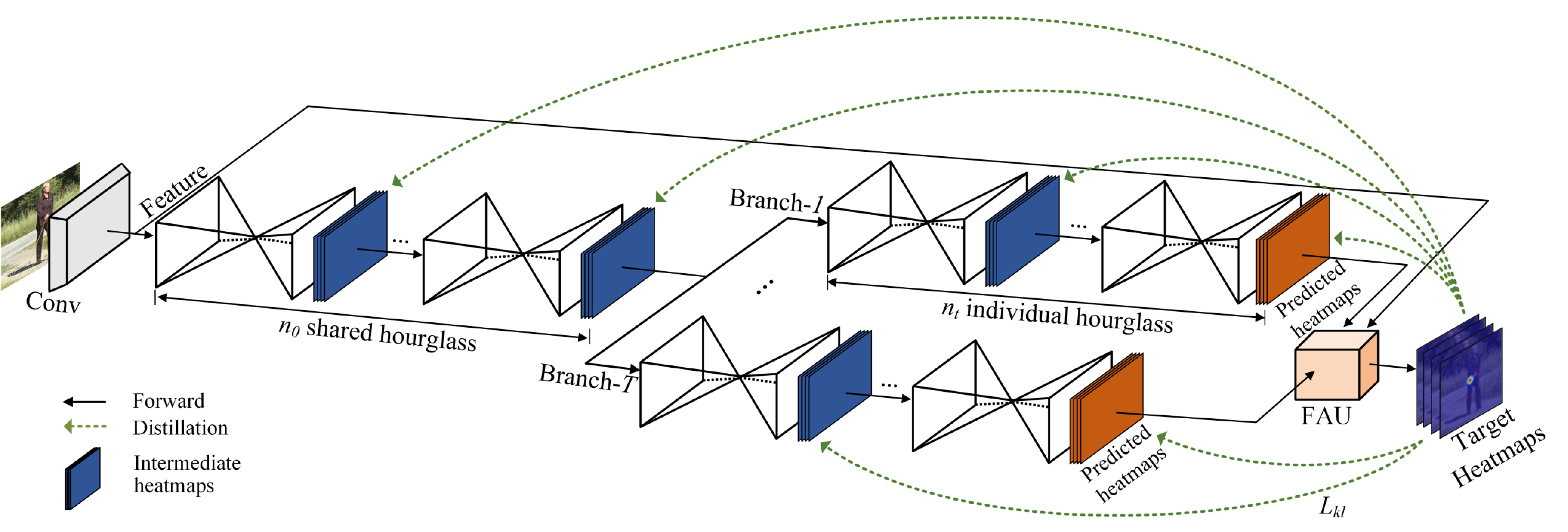} 
	\caption{
		An overview of the proposed Online Knowledge Distillation for Human Pose estimation (OKDHP). Each branch serves as an independent pose estimator. The FAU learns to ensemble all branches to establish a stronger teacher model. $L_{kl}$ denotes the KL divergence loss between intermediate heatmaps and ensemble heatmaps. We omit the conventional mean squared error loss $L_{mse}$ for simplicity.
	}
	\label{fig:okdhp}
\end{figure*}

\section{Related Work}

\textbf{Human Pose Estimation} 
Classical human pose estimation approaches mainly adopt the technique of pictorial structures~\cite{felzenszwalb2005pictorial,andriluka2009pictorial,johnson2010clustered,pishchulin2013poselet,pishchulin2013strong} and graphical models~\cite{sapp2010cascaded,sapp2013modec,chen2014articulated,cherian2014mixing}. With the rapid development of deep convolutional neural networks~\cite{krizhevsky2012imagenet, simonyan2014very, he2016deep}, approaches based on CNNs became popular in recent years~\cite{tompson2014joint,wei2016convolutional,chu2017multi,nie2018human,sun2019deep,artacho2020unipose}. DeepPose proposed by Toshev \textit{et al.}~\cite{toshev2014deeppose} was the first attempt to regress the coordinates of body parts directly and shows superior performance than classical approaches. Tompson \textit{et al}.~\cite{tompson2014joint} learned body structures by jointly optimize the convnets and graphical models. CPM~\cite{wei2016convolutional} incorporate convolutional networks into the pose machine framework for the task of human pose estimation and directly performs pose matching on the heatmaps. Newell \textit{et al.}~\cite{newell2016stacked} stacked several hourglass modules to iteratively refine the predictions. Intermediate supervision is also used to produce accurate intermediate heatmaps and prevent gradient vanishes. The hourglass module is highly related to conv-deconv architecture~\cite{long2015fully,ronneberger2015u}. Features in this module are first pooled down to a low resolution, then are upsampled and fused with high-resolution features. Chu \textit{et al.}~\cite{chu2017multi} try to incorporate hourglass networks with attention mechanisms to learn and infer contextual representations. Yang \textit{et al.}~\cite{yang2017learning} further improved its performance by using the pyramid residual model. 

In addition to heavy networks for highly accurate pose estimation, highly efficient pose estimation networks have also been studied to meet the needs of real applications. Cao \textit{et al.}~\cite{cao2017realtime} introduced a real-time estimation network with two branches where one branch generates the heatmap predictions, while the other one generates part affinity field, then a greedy algorithm is used to group the joints to the corresponding person. Kocabas \textit{et al.}~\cite{kocabas2018multiposenet} proposed pose residual network that takes as input keypoints and person detections then perform keypoints assignment. MultiPoseNet achieves similar accuracy to Mask-RCNN~\cite{he2017mask} while being at least 4x faster. Based on OpenPose~\cite{cao2017realtime}, ~\cite{osokin2018real} uses a MobileNet~\cite{howard2017mobilenets} as a backbone network and adopt lightweight refinement stage to reduce computational cost.

\textbf{Knowledge Distillation}
Originally introduced by Hinton \textit{et al.}~\cite{hinton2015distilling}, knowledge distillation transfers knowledge in the form of soft predictions from a large and computational expensive model to a single computational efficient model through a learning procedure. When training the target student model, this method makes full use of the extra supervisory signal provided by the soft output of the teacher model.
In FitNet~\cite{romero2014fitnets}, the student was forced to mimic the intermediate feature representations of the teacher. AT~\cite{zagoruyko2016paying} try to transfer attention map of the teacher to the student. Kim \textit{et al.}~\cite{kim2018paraphrasing} introduces the paraphraser and translator network to assist the knowledge transfer procedure. In FSP~\cite{yim2017gift}, the student mimics the teacher's flow matrices, which are calculated as the inner product between feature maps from two layers. Traditional distillation methods always start with a powerful and cumbersome teacher model and perform one-way knowledge transfer to a compact student model. Online knowledge distillation~\cite{zhang2018deep,zhu2018knowledge} simplifies the complex two-stage procedure by reducing the need for a pre-trained teacher model. ONE~\cite{zhu2018knowledge} builds a single multi-branch network and each branch learns from the ensemble results. Chen \textit{et al.}~\cite{chen2020online} introduces the two-level distillation framework and uses a self-attention mechanism to construct diverse peer networks. Li \textit{et al.}~\cite{li2020online} made a further improvement to such branch-based network by enhancing the branch diversity.
\vspace{-1em}

Knowledge distillation methods have been widely used in many vision tasks, including object detection~\cite{li2017mimicking,chen2017learning,deng2019relation}, line detection~\cite{hou2019learning}, semantic segmentation~\cite{Ye2019StudentBT,he2019knowledge,liu2019structured} and human pose estimation~\cite{zhang2019fast,nie2019dynamic,wang2019distill,weinzaepfel2020dope}. DOPE~\cite{weinzaepfel2020dope} proposes to distill the 2D and 3D poses from three independent body part expert models to the single whole-body pose detection model. Nie \textit{et al.}~\cite{nie2019dynamic} distill the pose kernels via leveraging temporal cues from the previous frame in a one-shot feed-forward manner. Wang \textit{et al.} distill the 3D pose knowledge from Non-Rigid Structure from Motion in weakly supervised learning. FPD~\cite{zhang2019fast} adopts the classical distillation approach and transfers the knowledge from an 8-Stack hourglass network to a lightweight 4-Stack hourglass network. Two shortcomings exist in the above work. A high-capacity teacher model is not always available and such complex two-stage learning will make the distillation inefficient.

\section{Methodology}

In this section, we first present a brief introduction to knowledge distillation, then we describe our proposed online knowledge distillation framework for efficient human pose estimation. Finally, we introduce the unbalanced version of our proposed OKDHP.

\subsection{Teacher-Student Learning}

Knowledge distillation~\cite{hinton2015distilling}, as one of the main model compression techniques~\cite{wu2016quantized,molchanov2016pruning}, follows the classic teacher-student learning paradigm. By treating a pre-trained heavy network as the teacher model, knowledge distillation aims to learn a lightweight student model, which is expected to master the expertise of the teacher, via transferring the knowledge from the teacher. Such distillation procedure can be formulated as:
\begin{equation}
L_{kd} = d(m_{stu},m_{tea}),
\end{equation}
where $d(\cdot)$ denotes the distance loss function, measuring the differences between two probability distributions. $m_{stu}$ and $m_{tea}$ represent the results generated by the student and the teacher, respectively.

With the task-specific supervised loss $L_{task}$, the whole loss function is given as:
\begin{equation}
L_{total}=L_{task}+\lambda L_{kd},
\end{equation}
where $\lambda$ is the hyperparameter for balancing the two loss terms.

The vanilla knowledge distillation is a two-stage procedure where a cumbersome teacher model is first trained and fixed, and then the knowledge is distilled to a compact student model. This process increases the training complexity, making the distillation process inefficient. 

\subsection{Online Human Pose Distillation}

To solve the problems in the vanilla distillation method, we propose an online knowledge distillation framework for efficient pose estimation. An overview of the proposed framework is illustrated in Fig.~\ref{fig:okdhp}. The proposed OKDHP architecture contains a multi-branch network as the main network and an FAU module for building the teacher online. We adopt the Hourglass network~\cite{newell2016stacked} (HG) as our basic building block in the proposed framework, which is the most common block used in many state-of-the-art works~\cite{chu2017multi,ke2018multi,li2020simple}. 

\subsubsection{The Main Network}

The main network is in the multi-branch architecture that consists of $T$ auxiliary homogeneous branches with the same network configuration (the same number of HG modules). That is, a total of $T$ pose estimators are aggregated in the main network, each of which shares the first $n_{0}$ HG modules and is treated as a student. For every $1\le t\le T$, branch-$t$ has $n_{t}$ individual HG modules. To make the method clearer, we firstly give the details of the OKDHP-balanced, where $n_{1}=n_{2}=...=n_{T}$.

Thus, given a single RGB image, human pose estimation estimates a heatmap for each human anatomical keypoint, which represents the keypoint locations as Gaussian peaks. To train the multi-branch main network, we minimize the mean squared error (MSE) between the predicted heatmaps $m_{pred}$ from each branch and the ground-truth heatmaps $m_{gt}$:
\begin{equation}
L_{mse} =\frac{1}{C}\sum_{t=0}^{T}\sum_{c=1}^C\|m_{gt}^{c}-m_{pred}^{c}(t)\|_{2}^{2},
\end{equation}
where $C$ denotes the total number of human keypoints i.e. heatmap channels and $T$ denotes the total number of network branches. $m_{pred}^{c}(t )$ is the predicted heatmap from branch-$t$ at the $c$-th channel. Note that our network is built by stacking multiple hourglass modules, the supervision is applied not only on the final output but also on the intermediate heatmaps from each HG module.

\begin{figure*}
	\centering
	\includegraphics[width=1\linewidth]{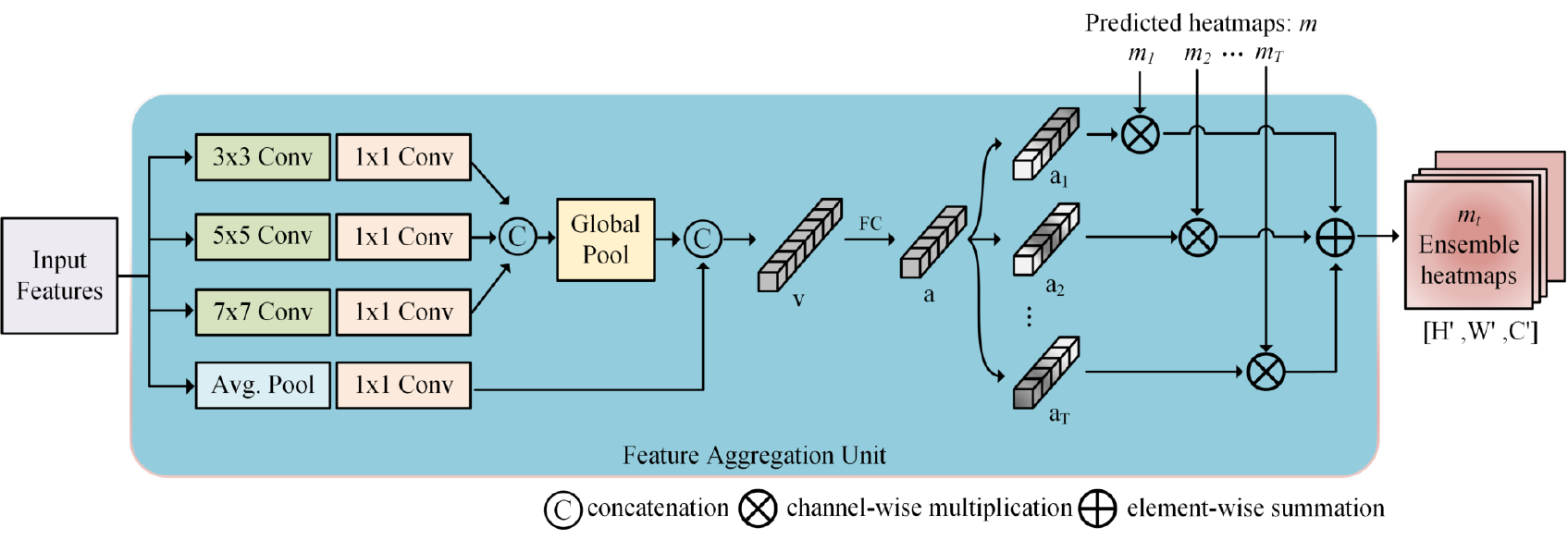}
	\caption{
		This module is proposed to effectively exploit the multi-scale information that can obtain target heatmaps with higher-quality. $m_{i}$ denotes the predicted heatmaps that come from $i$-th branch. The final ensemble heatmaps are obtained by the weighted sum of heatmaps of all individual branches. Note that all conv blocks in the network are composed of regular convolutions, Batch Normalization and ReLU activation functions in sequence. 
	}
	\label{fig:fau}
\end{figure*}

\subsubsection{The FAU Module}
The Feature Aggregation Unit (FAU) learns to combine all the predicted heatmaps from $T$ branches to establish a strong teacher model. The FAU module consists of multiple parallel transformations with different receptive fields, leveraging both local and global information to obtain accurate target heatmaps. The architecture of the proposed FAU is depicted in Fig.~\ref{fig:fau}. 

Previous image classification work~\cite{zhu2018knowledge} uses a simple conv block as the gate module to generate an importance score for each branch. But a simple conv block cannot effectively capture the contextual representation due to the body scale variation problem that exists in the natural scene. The multi-scale information is required to handle this problem.
In this work, we focus on effectively capturing the multi-scale information to generate the target heatmaps with higher-quality. Inspired by the previous works~\cite{li2019selective,gao2019res2net}, we propose the FAU which is composed of multiple parallel transformations with different receptive fields.
As shown in Fig.~\ref{fig:fau}, multiple conv blocks in FAU have different receptive fields and are arranged in parallel. We take as input the features after the main network Conv block which contains more original information. 
The convolution operation starts with a small kernel size of 3, then consistently increases in the following branches (size of 3,5,7). In our network, additional 1$\times$1 convolutions are mainly used as dimensional reduction methods to save computational resources. We further combine the average pooling of original inputs for richer representations. Then we concatenate all the splits and obtain the intermediate vector \textbf{v}, denoted as:
\begin{equation}
\textbf{v}= [v_{avg},g([v_{conv3},v_{conv5},v_{conv7}])],
\end{equation}
where $g(\cdot)$ denotes the global pooling function. $v_{conv3}$, $v_{conv5}$, $v_{conv7}$ denote the results from the conv path with kernel size 3, 5 and 7, respectively. $v_{avg}$ denotes the results from the average pooling path.

For any input feature maps, this configuration creates multi-scale features with each conv path that are aggregated to capture richer information for both local and global fields. We pass the intermediate vector $\textbf{v}$ through the fully connected layer $FC$ to fuse the information from different paths. Then, a channel-wise softmax operator is applied to obtain the soft attention vectors $a_{t,c}$. 
In the case of three network branches, for $c$-th heatmap we have
\begin{equation}
a_{1,c}+a_{2,c}+a_{3,c}=1,
\end{equation}
where c=1, 2, ..., $C$.
Finally, we fuse predictions from multiple branches via an element-wise summation to obtain the weighted target heatmaps $\textbf{m}_{tar}$:
\begin{equation}
\textbf{m}_{tar} = \sum_{t=1}^{T} \textbf{a}^{t}\otimes\textbf{m}_{s}^{t},
\end{equation}
where $\textbf{a}^{t}=[a_{1},a_{2},..., a_{c}]^{t}$, $\textbf{m}_{s}^{t}=[m_{s}^{1},m_{s}^{2},...,m_{s}^{c}]^{t}$ and $\textbf{m}_{t}\in\mathbb{R}^{H'\times W'\times C}$. Here, $\textbf{a}^{t}$ is the weight for the $t$-th branch, $\textbf{m}_{s}^{t}$ is the heatmaps generated by the \textit{t}-th branch and $\otimes$ refers to the channel-wise multiplication between $\textbf{a}^{t}$ and $\textbf{m}_{s}^{t}$. Our experiments (see Section 4.4) prove that the weights generated by FAU can achieve better distillation performance.

\subsubsection{Pixel-wise Distillation}
A proper distillation loss function is critical to the whole training procedure. Since pixel values on the heatmaps indicate the probabilities of pixels that belong to the keypoint. We align the heatmap generated by the student model with the target heatmaps. The target heatmaps obtained through the FAU play the role of a teacher model to teach each branch model (student) in our method. To transfer the pose structural knowledge, the pixel-wise Kullback-Leibler (KL) divergence loss is utilized to minimize the divergence between the heatmaps of the teacher model and the student model as follows:
\begin{equation}
L_{kl} = \frac{1}{W'\times H'}\sum_{i\in M}\sum_{t=0}^{T}KL(q_{tar}^{i} , q(t)_{s}^{i}),
\end{equation}
where $W'$ and $H'$ represent the heatmaps' width and height. $q_{tar}^{i}$ and $q(t)_{s}^{i}$ denote the probabilities of the \textit{i}-th pixel from the heatmap generated by the teacher model and the student model, respectively. $M=\{1,2,...,W'\times H'\}$ denotes all the pixels.

\textbf{Overall}
To get a better understanding of our method, we describe the whole training procedure in Algorithm~\ref{algo:demo}.
For the proposed online human pose distillation method, the whole objective function consists of a conventional mean squared error loss $L_{mse}$ for pose estimation and another loss term $L_{kl}$ for online knowledge distillation:
\begin{equation}
L_{total} = \alpha L_{mse}+\beta L_{kl}.
\end{equation}
where $\alpha$ and $\gamma$ are the hyperparameters to balance these two losses. 

\begin{algorithm}[t]
	\caption{Online Human Pose Distillation}
	\hspace*{0.02in} {\bf Input:}
	Labelled Training dataset $D$; Training Epoch Number $\epsilon$; Branch Number $T$; Network Structure $S\in\{S_{balance}, S_{unbalance}\}$ \\
	\hspace*{0.02in} {\bf Output:}
	Trained target pose estimate network $\theta^{1}$ and other auxiliary estimators $\{\theta^{i}\}^{T}_{i=2}$ \\
	\hspace*{0.02in} {\bf Initialize:}
	Epoch e=1; Randomly initialize $\{\theta^{i}\}^{T}_{i=1}$
	
	\begin{algorithmic}[1]
		\While {e $\leq$ $\epsilon$}
		\State Compute the heatmap predictions of all branches $\{\theta^{i}\}^{T}_{i=1}$ according to $S$;
		\State Compute the target heatmaps $m_{t}$  through FAU;
		\State Compute the MSE loss $L_{mse}$;
		\State Compute the distillation loss $L_{kl}$;
		\State Compute the total loss function;
		\State Update the model parameters $\{\theta^{i}\}^{T}_{i=1}$;
		\State e=e+1;
		\EndWhile
		\State \textbf{end while}
	\end{algorithmic}
	\hspace*{0.02in} {\bf Model deployment:}
	Use target pose estimator $\theta^{1}$;
	\label{algo:demo}
\end{algorithm}

\begin{table*}
	\begin{center}
		{
			\begin{tabular}{ccccccccc}
				\hline
				Method          & Head & Sho. & Elb. & Wri. & Hip & Knee & Ank. & Mean \\
				\hline
				\hline
				Newell \textit{et al.}(HG) [ECCV'16]~\cite{newell2016stacked} & 98.2 & 96.3 & 91.2 & 87.1 & 90.1 & 87.4 & 83.6 & \textbf{90.9}\\
				Ning \textit{et al.} [TMM'17~\cite{ning2017knowledge}] & 98.1 & 96.3 & 92.2 & 87.8 & 90.6 & 87.6 & 82.7 & 91.2 \\
				Chu \textit{et al.} [CVPR'17]~\cite{chu2017multi} & 98.5 & 96.3 & 91.9 & 88.1 & 90.6 & 88.0 & 85.0 & 91.5 \\
				Chen \textit{et al.} [ICCV'17]~\cite{chen2017adversarial} & 98.1 & 96.5 & 92.5 & 88.5	& 90.2 & 89.6 & 86.0 & 91.9 \\
				Yang \textit{et al.} [ICCV'17]~\cite{yang2017learning} & 98.5 & 96.7 & 92.5 & 88.7 & 91.1 & 88.6 & 86.0 & 92.0 \\
				Ke \textit{et al.} [ECCV'18]~\cite{ke2018multi} & 98.5 & 96.8 & 92.7 & 88.4	& 90.6 & 89.3 &	86.3 & 92.1 \\
				Tang \textit{et al.} [ECCV'18]~\cite{tang2018deeply} & 98.4 & 96.9 & 92.6 & 88.7 & 91.8 & 89.4 & 86.2 & 92.3 \\
				\hline
				FPD [CVPR'19]~\cite{zhang2019fast} & 98.3 & 96.4 & 91.5 & 87.4 & 90.9 & 87.1 & 83.7 & 91.1 \\
				\hline
				OKDHP 		  & 98.2 & 96.6 & 92.3 & 88.0 & 91.0 & 88.5 & 84.5 & \textbf{91.7} \\
				\hline
			\end{tabular}
		}
	\end{center}
	\caption{Evaluation of our proposed OKDHP on MPII testing set (PCKh@0.5).}
	\label{table:compare_sota}
\end{table*}

\subsection{Unbalanced Architecture}
To achieve a better distillation performance, a stronger teacher model is required. But in our balanced architecture, the teacher is fixed once we set up the target network. Here, the unbalanced variant is introduced to customize the student model with different compression rates, as shown in Fig.~\ref{fig:okdhpv2}(c). For an unbalanced OKDHP architecture, each branch have different numbers of HG modules, where $n_{1}\neq n_{2}\neq...\neq n_{T}$. For example, for a 3-branch unbalanced network, if a 4-Stack HG network is required for final deployment, the other two branches can be set as a 6-stack and an 8-Stack network. Compared with the balanced structure, a better teaching performance can be achieved by utilizing the stronger representation ability of a larger network. Furthermore, the other benefit is that we can simultaneously obtain three different networks with comparable performances in one training procedure. This kind of network setting can be customized to other settings according to the actual needs. We demonstrate the effectiveness of unbalanced structure in Table~\ref{table:compare} and present the detailed results in Table~\ref{table:detail}.

\section{Experiments}

To validate the effectiveness of our proposed method, we conduct several experiments on two popular human pose datasets, MPII~\cite{andriluka14cvpr} and COCO~\cite{lin2014microsoft}.

\subsection{Implementation Details}

\textbf{Datasets} The MPII dataset includes approximately 25K images containing over 40K subjects with annotated body joints, where 29K subjects are used for training and 11K subjects are used for testing. The images were collected using an established taxonomy of everyday human activities from YouTube videos. We adopted the same train/valid/test split as in ~\cite{zhang2019fast}. Each person instance in MPII has 16 labeled joints. 

The COCO keypoint dataset~\cite{lin2014microsoft} presents naturally challenging imagery data with various poses. It contains more than 200k images and 250k person instances labeled with keypoints. In test, we follow the commonly used train/val/ test split. Each person instance is labeled with 17 joints.

\textbf{Training details} We implement all the methods in PyTorch~\cite{paszke2019pytorch}. For MPII, we resize the cropped images to 256$\times$256 in pixels. Then we randomly augment the data with rotation degrees in [-$30^{\circ}$, $30^{\circ}$], scaling with factors in [0.75, 1.25] and horizontal flip. For COCO, we resize the cropped image to 256$\times$192 in pixels. Then we apply random horizontal flip, random rotation with degrees in [-$40^{\circ}$, $40^{\circ}$] and random scale with factors in [0.7, 1.3]. We follow the standard data processing scheme for all images as in \cite{zhang2019fast}. Adam~\cite{kingma2014adam} is used as the optimizer and we set the initial learning rate to 2.5e-4, weight decay to 1e-4. The learning rate is divided by 10 at 90 and 120 of the total 150 training epochs. We usually set $\alpha$=1 and $\beta$=2 in Eqn.8. For the network architecture, we set the number of shared HG modules to half the number of total stacks. In the case of a 4-Stack OKDHP network, we have two shared HG modules and two individual HG modules for each branch. We set branch size to 3 as default. We use the OKDHP-balance architecture as our default scheme in the following experiments unless we specified. We adopt the official hourglass configurations as our baseline method in all experiments. Code will be released soon.

\begin{table}
	\begin{center}
		\resizebox{1\linewidth}{!}
		{
			\begin{tabular}{c|c|cccccccc}
				\hline
				OKDHP      & Network                         & Head & Sho. & Elbo. & Wri. & Hip  & Knee & Ank. & PCKh@0.5   \\
				\hline\hline
				$\times$   &\multirow{2}*{2-Stack HG}& 96.7 & 95.3 & 89.2  & 84.0 & 87.8 & 83.9 & 79.5 & 88.6  \\
				\checkmark & ~                              & 96.7 & \textbf{95.4} & \textbf{89.9} & \textbf{84.1} & \textbf{89.0} & \textbf{84.7} & \textbf{81.1} & \textbf{89.2}  \\
				\hline
				$\times$   &\multirow{2}*{4-Stack HG} & 96.7 & 95.6 & 89.7 & 84.5 & 88.6 & 84.3 & 80.9 & 89.2   \\
				\checkmark & ~                               & \textbf{97.0} & \textbf{96.1} & \textbf{90.8} & \textbf{85.9} & \textbf{89.5} & \textbf{85.4} & \textbf{81.6} & \textbf{90.0}  \\
				\hline
				$\times$   &\multirow{2}*{8-Stack HG} & 96.9 & 95.9 & 90.6 & 86.0 & 89.8 & 86.0 & 82.5 & 90.2   \\
				\checkmark & ~                        		 & \textbf{97.3} & \textbf{96.1} & \textbf{91.2} & \textbf{86.8} & \textbf{89.9}& \textbf{86.9} & \textbf{83.1} & \textbf{90.6}  \\
				\hline
			\end{tabular}
		}
	\end{center}
	\caption{PCKh@0.5 score of our proposed OKDHP on MPII validation set.}
	\label{table:mpii}
\end{table}

\begin{table}
	\begin{center}
		{
			\begin{tabular}{c|c|c}
				\hline
				Method          & PCKh@0.5 & TrainCost\\
				\hline
				\hline
				Baseline 		& 89.2 & 14 \\
				\hline
				FPD      		& 89.7 & 66\\
				\hline
				OKDHP-Balance   & 90.0 & 47\\
				\hline
				OKDHP-Unbalance & 90.2 & 64\\
				\hline
			\end{tabular}
		}
	\end{center}
	\caption{Comparison with different distillation methods based on the 4-Stack hourglass network on MPII validation set. TrainCost: Training cost in the unit of GFLOPS.}
	\label{table:compare}
\end{table}

\begin{table*}[h]
	\begin{center}
		{
			\begin{tabular}{c|c|cccccccccc}
				\hline
				OKDHP       & Network    				      & $AP$ & $AP^{50}$ & $AP^{75}$ & $AP^{M}$ & $AP^{L}$ & $AR$ & $AR^{50}$ & $AR^{75}$ & $AR^{M}$ & $AR^{L}$\\
				\hline\hline
				$\times$    &\multirow{2}*{2-Stack HG} &  71.7 & 90.5 & 78.4 & 69.0 & 75.8 & 74.6 & 91.9 & 80.6 & 71.6 & 79.2 \\
				\checkmark	& ~          				      &  \textbf{72.8} & \textbf{91.5} & \textbf{79.5} & \textbf{69.9} & \textbf{77.1} & \textbf{75.6} & \textbf{92.5} & \textbf{81.5} & \textbf{72.5} & \textbf{80.3} \\
				\hline
				$\times$  	&\multirow{2}*{4-Stack HG} &  73.6 & 91.6 & 80.6 &
				70.8 & 78.0 & 76.5 & 92.6 & 82.8 & 73.5 & 81.2\\
				\checkmark 	& ~							      &  \textbf{74.8} & \textbf{92.5} & \textbf{81.6} & \textbf{72.1} & \textbf{78.5} & \textbf{77.4} & \textbf{93.1} & \textbf{83.6} & \textbf{74.5} & \textbf{81.9} \\
				\hline
				$\times$  	&\multirow{2}*{8-Stack HG} & 75.3 & 91.6 & 82.6 & 73.0 & 79.1 & 78.0 & 92.9 & 84.0 & 75.2 & 82.3 \\
				\checkmark & ~							      & \textbf{76.2} & \textbf{92.6} & \textbf{83.7} & \textbf{73.4} & \textbf{80.2} & \textbf{78.8} & \textbf{93.6} & \textbf{85.2} & \textbf{75.9} & \textbf{83.3} \\
				\hline
			\end{tabular}
		}
	\end{center}
	\caption{Evaluation of our proposed OKDHP on COCO val2017 dataset.}
	\label{table:coco}
\end{table*}

\begin{table}
	\begin{center}
		{
			\begin{tabular}{c|c|c}
				\hline
				OKDHP-Unbalance & Network & PCKh@0.5 \\
				\hline
				\hline
				Branch-1 (Target)	   & 4-Stack HG & 90.2 \\
				\hline
				Branch-2 (Auxiliary)   & 6-Stack HG & 90.3 \\
				\hline
				Branch-3 (Auxiliary)   & 8-Stack HG & 90.5 \\
				\hline			
			\end{tabular}
		}
	\end{center}
	\caption{Detailed results of the 3-branch OKDHP-Unbalance (two shared HG modules) network on MPII validation set.}
	\label{table:detail}
\end{table}

\textbf{Evaluation Metric} We use the standard Percentage of Correct Keypoint (PCK) metric which reports the percentage of correct keypoints lies within a normalized distance of ground truth. We use PCKh@0.5 for the MPII dataset, which refers to a threshold of 50\% of the head diameter. For COCO, we use Object Keypoints Similarity (OKS) as our evaluation metric, which defines the similarity between different human poses.

\subsection{Results on MPII dataset}
\begin{figure}
	\centering
	\includegraphics[width=1\linewidth]{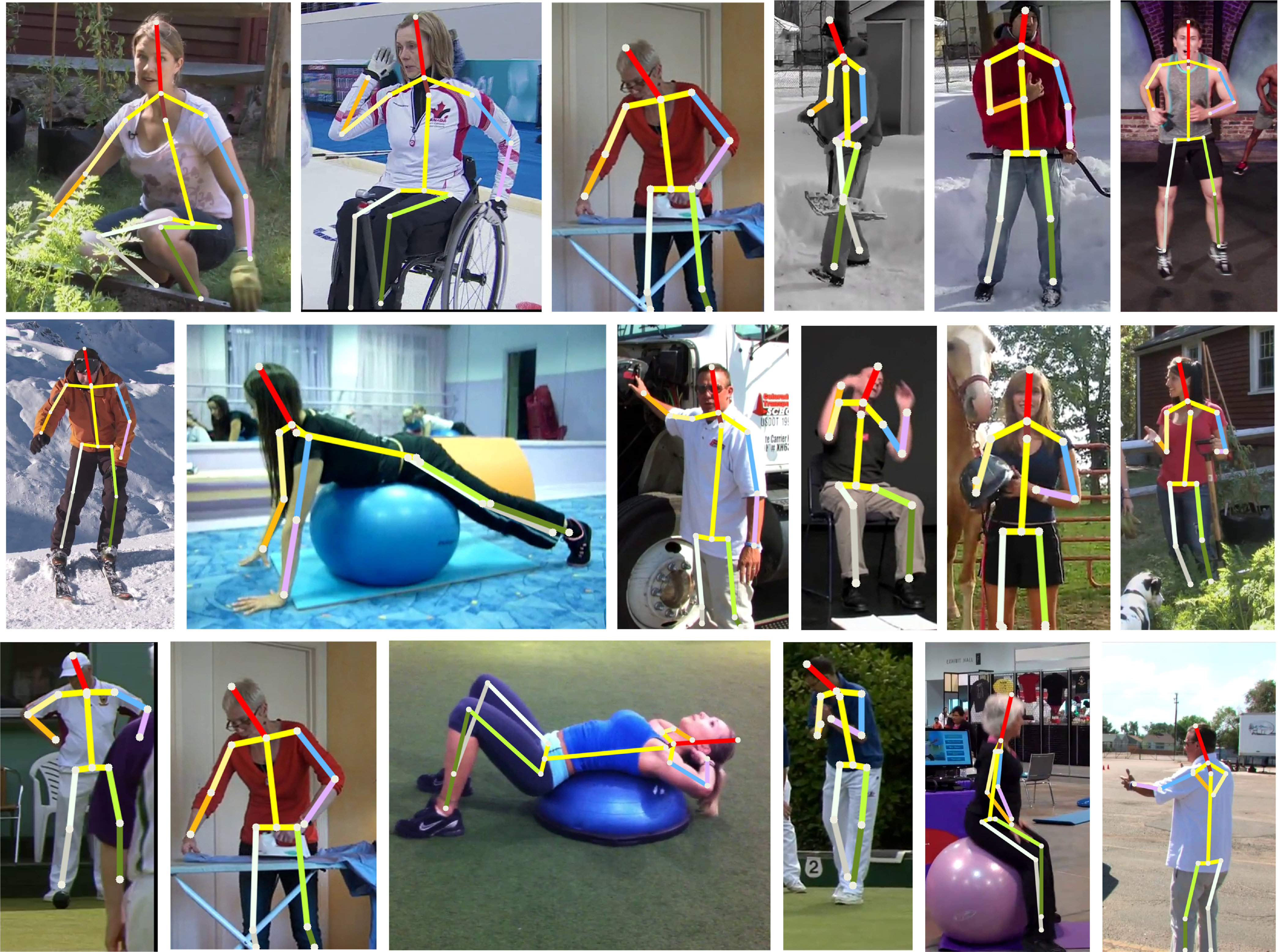}
	\caption{
		Visualized results on MPII dataset.
	}
	\label{fig:visualization}
\end{figure}

We evaluate our method on the MPII dataset. Table~\ref{table:compare_sota} compares the PCKh@0.5 accuracy results of state-of-the-art methods and our proposed OKDHP on the MPII testing set. Table~\ref{table:mpii} reports the comparison of three varying-capacity networks trained by the conventional method and our proposed OKDHP. We can clearly observe that all networks benefit from our OKDHP training method, particularly for small networks achieving large performance gains. Specifically, our method improves various baseline networks ranging from 0.3 to 0.8. Considering that the performance of many state-of-the-art pose estimation networks, improves from 0.1\% to 0.3\% in PCKh scores, our performance is in fact, significant as compared to prior works. A 2-Stack HG network trained with OKDHP achieves similar performance with the original 4-Stack HG network but it only needs half the number of HG modules. In contrast to the conventional distillation method, a large pre-trained teacher is not necessary. We provide the visualized pose results in Fig.~\ref{fig:visualization}.

We compare our method with previous state-of-the-art distillation work FPD~\cite{zhang2019fast} and demonstrate the performance comparison of our proposed balanced and unbalanced structure in Table~\ref{table:compare}. The teacher network is an 8-Stack HG network with 90.2 PCKh@0.5 scores for FPD. We can clearly observe that both OKDHP balanced and unbalanced architectures outperforms the FPD method, validating the performance advantage of our method. The unbalanced structure outperforms the balanced structure by 0.2\% but with 36\%(17/47) FLOPS increase. OKDHP-Balance takes the least training cost, proves that our method is the most effective pose distillation approach.

We provide the detailed results of our proposed OKDHP-Unbalance architecture in Table~\ref{table:detail}. Three branches exists in our network. The first one is the target 4-Stack hourglass network for deployment. The other two branches plays the auxiliary roles, helping the target branch to achieve better performance.

\subsection{Results on COCO dataset}

Table~\ref{table:coco} shows the results of the baseline method and our proposed OKDHP on the MS COCO keypoint dataset. In the test, a two-stage top-down paradigm is applied, same as in ~\cite{xiao2018simple,chen2018cascaded,zhang2020distribution}. We use a detector with detection AP 56.4 for the person category on COCO val2017. From this table, we can observe that OKDHP method yields a more generalizable model compared to independent learning. 
This demonstrates that our method can still be applied to the large-scale dataset effectively.

\subsection{Ablation Study}

\begin{table}
	\begin{center}
		\resizebox{1\linewidth}{!}
		{
			\begin{tabular}{l|cccccccc}
				\hline
				Loss & Head & Sho. & Elbo. & Wri. & Hip & Knee & Ank. & PCK \\
				\hline\hline 
				MSE & 96.9 & 95.8 & 89.9 & 84.8 & 89.3 & 84.9 & 81.5 & 89.5 \\
				\hline
				Ours & \textbf{97.0} & \textbf{96.1} & \textbf{90.8} & \textbf{85.9} & \textbf{89.5} & \textbf{85.4} & \textbf{81.6} & \textbf{90.0} \\
				\hline
			\end{tabular}
		}
	\end{center}
	\caption{Comparisons of different distillation loss functions on MPII validation set (PCKh@0.5).}
	\label{table:loss}	
\end{table}

\textbf{Loss Function}
The distillation loss function plays a critical role in the whole learning procedure. We compared the performance of different distillation loss functions as shown in Table~\ref{table:loss}. FPD uses mean squared error (MSE) loss as the distillation term which is the same as the task-specific supervised term. We test the MSE loss in our proposed framework. In Table~\ref{table:loss}, this result shows that our proposed pixel-wise KL divergence is a better choice in comparison to the MSE. Our method can effectively distill the pose structural knowledge to enhance distillation performance.

\textbf{Branch Size} We evaluate the impact of branch size on the performance of our branch-based online pose distillation framework. Table~\ref{table:branch} shows the performance on MPII validation set with varying branch sizes ranging from 2 to 5. We omit the case when branch size $n=1$ since one branch cannot form the ensemble result. The baseline method denotes the vanilla 2-Stack networks without any modifications. We can clearly observe that OKDHP scales well with more branches and a 2-Stack OKDHP can be further improved if a larger branch size is allowed during training. 

\begin{table}[h]
	\begin{center}
		{
			\begin{tabular}{c|cc}
				\hline
				Branch Size & PCKh@0.5 & \#Params \\
				\hline\hline
				Baseline & 88.6 & 13.0M\\
				\hline
				2 & 89.2 & 15.5M\\
				3 & 89.2 & 18.6M\\
				4 & 89.3 & 21.7M\\
				5 & \textbf{89.4} & 24.7M\\
				\hline
			\end{tabular}
		}
	\end{center}
	\caption{Impact of branch size for a 2-Stack OKDHP framework on MPII validation set (PCKh@0.5).}
	\label{table:branch}
\end{table}

\textbf{Individual HG Number}
We set $n_{s}=2$ and $n_{i}=2$ for a 4-Stack OKDHP network in the main experiment, indicating that we have two shared HG modules and two individual HG modules for each branch. Table~\ref{table:individual_number} demonstrated the impact of the individual HG numbers on MPII validation set. From this table, we can see that if very few HG modules are shared, the performance will quickly drop. This brings the concept of branch diversity for such branch-based networks as mentioned in ~\cite{chen2020online,li2020online}. Diversity will be hurt with the reducing number of individual HG modules, which will limit the effectiveness of within-group knowledge transfer. We usually set the number of shared and individual modules to half the number of total stacks to achieve the accuracy-efficiency trade-offs.

\begin{table}[h]
	\begin{center}
		{
			\begin{tabular}{c|ccccccc}
				\hline
				Individual HG numbers  & 1 & 2 & 3 & 4 \\
				\hline\hline 
				PCKh@0.5 			   & 89.8 & 90.0 & 90.1 & \textbf{90.2} \\
				\hline
				FLOPs				   & 41G  & 47G  & 53G  & 59G \\
				\hline
			\end{tabular}
		}
	\end{center}
	\caption{Impact of the number of individual HG modules for a 4-Stack OKDHP network on MPII validation set (PCKh@0.5).}
	\label{table:individual_number}
\end{table}

\textbf{Sensitivity to Hyperparameter} Table~\ref{table:sensitivity} demonstrates how the performance of our proposed framework is affected by the choice of hyperparameter $\beta$ in Eqn.8. From this table, we can see that our method still has robust performance against varying $\beta$ values ranging from 0.5 to 5.

\begin{table}[h]
	\begin{center}
		\resizebox{1\linewidth}{!}
		{
			\begin{tabular}{c|ccccccc}
				\hline
				$\beta$  & 0.5 & 1 & 2 & 3 & 4 & 5 \\
				\hline\hline 
				PCKh@0.5 & 89.24 & 89.20 & \textbf{89.28} & 89.19 & 89.20 & 89.18 \\
				\hline
			\end{tabular}
		}
	\end{center}
	\caption{Sensitivity to $\beta$ for a 2-Stack OKDHP network on MPII validation set (PCKh@0.5).}
	\label{table:sensitivity}
\end{table}

\textbf{FAU} 
The goal of FAU is to generate accurate target heatmaps by weighted ensemble heatmaps from all auxiliary branches. To evaluate the effectiveness of our proposed FAU, we conduct various ablation studies on MPII validation set based on a 4-Stack HG network as shown in Table~\ref{table:fau}. We compare the performance of the following experiments. 
(1) Baseline: A vanilla 4-Stack HG network without any modification.
(2) Mean: A simple average is applied to aggregate the heatmaps of all branches.
(3) Gate: A simple attention module used in ONE~\cite{zhu2018knowledge} for the classification task. It was initially proposed for image classification. We reimplement this module so that it can be directly used in pose estimation network.
(4) FAU: Our proposed module. We can see that FAU outperforms Mean and Gate by 0.3\% and 0.2\%, respectively. This confirms the usefulness of the FAU.

\begin{table}[h]
	\begin{center}
		{
			\begin{tabular}{cccc}
				\hline
				Baseline   & Mean & Gate & FAU \\
				\hline\hline
				89.2       & 89.7 & 89.8 & \textbf{90.0} \\
				\hline
			\end{tabular}
		}
	\end{center}
	\caption{Ablation study on FAU module for a 4-Stack HG network on MPII validation set (PCKh@0.5).}
	\label{table:fau}
\end{table}

\section{Conclusion}

In this paper, we propose a novel Online Knowledge Distillation framework by distilling Human Pose structure knowledge (OKDHP) in the one-stage manner. A network with multiple branches is utilized in the framework, where each branch is an independent pose estimator and is regarded as the student. The students from multiple branches are integrated into one teacher by the FAU module, which then optimizes the student branches in reverse.
With OKDHP, the efficiency is significantly enhanced with reduced distillation complexity and improved model performance. Besides, the unbalanced OKDHP scheme is also introduced to enable the customization of the target network with different compression rates. Experiments have validated the effectiveness of our proposed OKDHP on two popular benchmark datasets.

\noindent\textbf{Acknowledgement}
This work is supported by National Key Research and Development Project of China (Grant No.2017YFB1002803) and National Natural Science Foundation of China (Grant No.62072150).

{\small
\bibliographystyle{ieee_fullname}
\bibliography{egbib}
}

\section*{Supplementary Material}

\subsection*{Additional Experiments}

We further extend our proposed OKDHP to a popular light-weight and real-time pose estimation method called MobilePose\footnote{https://github.com/YuliangXiu/MobilePose-pytorch} with three popular backbone networks (ResNet-18, MobileNet-V2, ShuffleNet-V2). Table~\ref{table:compare2_real_time}  reports the comparison of baseline method and our OKDHP.
The MobilePose network can be roughly divided into two parts: the encoder (i.e. backbone network) and the decoder. The encoder extracts features from the input image. The decoder upsamples the features generated by the encoder, and obtains the keypoint heatmaps. In our proposed OKDHP, we share the encoder network and build multiple parallel decoder branches. The FAU module remains the same.

From Table~\ref{table:compare2_real_time}, we can observe that all networks benefit from our proposed OKDHP, particularly for small backbone networks achieving larger performance gains.

\begin{table}[h]
	\begin{center}
		\resizebox{1\linewidth}{!}
		{
			\begin{tabular}{cccccccccc}
				\hline
				OKDHP & Backbone    & Head & Sho. & Elb. & Wri. & Hip & Knee & Ank. & Mean \\
				\hline
				\hline
				$\times$ & \multirow{2}*{ResNet18}    & 95.8 & 93.9 & 84.9 & 78.3 & 85.7 & 79.4 & 74.5 & 85.4 \\
				\checkmark & ~ & \textbf{96.4} & \textbf{94.5} & \textbf{85.9} & \textbf{79.2} & \textbf{86.4} & \textbf{81.2} & \textbf{76.4} & \textbf{86.4} \\
				\hline
				$\times$ & \multirow{2}*{MobileNetV2} & 95.2 & 92.8 & 82.6 & 75.4 & 84.3 & 76.2 & 70.4 & 83.3 \\
				\checkmark & ~ & \textbf{95.5} & \textbf{94.2} & \textbf{84.8} & \textbf{78.1} & \textbf{85.9} & \textbf{78.7} & \textbf{73.6} & \textbf{85.1}\\
				\hline
				$\times$ & \multirow{2}*{ShuffleNetV2} & 94.8 & 91.2 & 79.4 & 70.9 & 82.2 &	71.8 & 66.5 & 80.5 \\
				\checkmark & ~ & \textbf{94.9} & \textbf{92.9} & \textbf{82.2} & \textbf{74.7} & \textbf{84.1} &	\textbf{76.0} & \textbf{70.0} & \textbf{83.0}\\
				\hline
			\end{tabular}
		}
	\end{center}
	\caption{PCKh@0.5 score on the MPII validation set.}
	\label{table:compare2_real_time}
\end{table}

\subsection*{Failure Case Analysis}

\begin{figure}[h]
	\small
	\centering
	\includegraphics[width=1\linewidth]{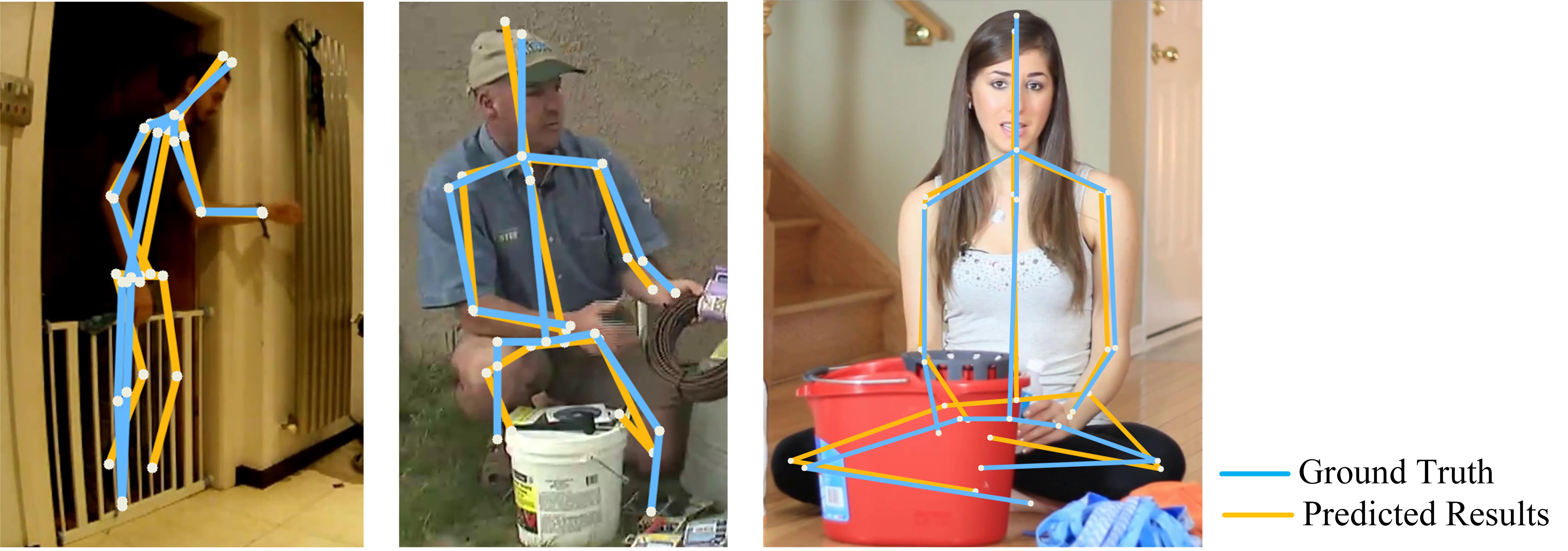}
	\caption{
		Visualized results of failed cases.
	}
	\label{fig:failuer_cases}
\end{figure}

The visualization of failed cases is shown in Fig.~\ref{fig:failuer_cases}. When the human body is occluded or the ambient light is insufficient, the predictions are still biased. Our proposed OKDHP method is a multi-branch online knowledge distillation framework based on a human pose estimation model. Although the model performance has been significantly improved after distillation, it is stilled limited by the inherited factors of the model itself.

\end{document}